# A Combinatorial Algebraic Approach for the Identifiability of Low-Rank Matrix Completion

**Franz Király**  FRANZ.J.KIRALY@TU-BERLIN.DE

Berlin Institute of Technology (TU Berlin), Machine Learning Group, Franklinstr. 28/29, 10587 Berlin, Germany, and FU Berlin, Discrete Geometry Group, Arnimallee 2, 14195 Berlin

**Ryota Tomioka**  TOMIOKA@MIST.I.U-TOKYO.AC.JP

University of Tokyo, Information-Theoretic Machine Learning and Data Mining Group, 7-3-1 Hongo, Bunkyo-ku, Tokyo 113-8656, Japan

## Abstract

In this paper, we review the problem of matrix completion and expose its intimate relations with algebraic geometry, combinatorics and graph theory. We present the first necessary and sufficient combinatorial conditions for matrices of arbitrary rank to be identifiable from a set of matrix entries, yielding theoretical constraints and new algorithms for the problem of matrix completion. We conclude by algorithmically evaluating the tightness of the given conditions and algorithms for practically relevant matrix sizes, showing that the algebraic-combinatorial approach can lead to improvements over state-of-the-art matrix completion methods.

## 1. Introduction

Reconstruction of a low-rank matrix from partial measurements arises naturally in many practically relevant problems, such as imputation of missing features, multi-task learning (Argyriou et al., 2008), transductive learning (Goldberg et al., 2010), as well as collaborative filtering (Srebro et al., 2005).

Following the success of the nuclear norm heuristic for matrix completion (Srebro et al., 2005; Candes & Recht, 2009), considerable effort has been devoted to understand the performance of such a procedure.

Previous studies can be classified by the assumptions about the sampling procedure and the underlying low-rank matrix. (Candes & Recht, 2009) analyzes the noiseless setting, and have shown under uniform sampling that incoherent low-rank matrices can be recovered with large probability. (Salakhutdinov & Srebro, 2010) considered the more realistic setting where the rows/columns are non-uniformly sampled. (Negahban & Wainwright, 2010) showed under the same row/column weighted sampling that non-spiky low-rank matrices can be recovered with large probability. (Foygel & Srebro, 2011) have shown under uniform sampling that the max-norm heuristic (Srebro & Shraibman, 2005) can achieve superior reconstruction guarantee without the non-spikiness assumption on the underlying low-rank matrix.

All the above theoretical guarantees are built on some assumption on the sampling procedure, e.g., uniform sampling. Nevertheless, in a practical setting, we always know which entries we can observe and which entries we cannot (the so-called mask). One may ask if we could obtain a stronger theoretical guarantee (of success or failure) conditioned on the mask we have.

On the other hand, all the above theories are also based on some assumptions on the underlying low-rank matrix, which are usually uncheckable. Although it is clear that we cannot recover arbitrary low-rank matrices (see Candes & Recht, 2009), one may ask if we could build a theory for matrix completion for almost all matrices, depending only on the mask.

Until now, all Machine Learning approaches have been mostly agnostic to the deep connections between matrix completion and the fields of combinatorics and algebraic geometry. Indeed there are at least two existing independent strains of work related to matrix completion outside the Machine Learning community: the first is rigidity theory, a branch of combinatorics concerned with realizing partially known distance ma-

# A Combinatorial Algebraic Approach for the Identifiability of Low-Rank Matrix Completion

trices which is subtly different from matrix completion, but is closely related, see (Singer & Cucuringu, 2010) for an overview. The second strain, in algebraic geometry, is the study of determinantal varieties. A determinantal variety is a set of low rank matrices with possible additional properties, considered as a manifold. The results closest to matrix completion in pure algebraic geometry are on sections of determinantal varieties (Giusti & Merle, 1982; Gonciulea & Miller, 2000; Boocher, 2011). Moreover, determinantal varieties frequently expose combinatorial structure which is studied in field of combinatorial algebra. Connections between combinatorics and algebra have already been observed in rigidity theory, crucially contributing to recent developments. We will show that matrix completion is also related naturally to both algebra and combinatorics.

We believe that using and combining these disjoint techniques may be highly beneficial for the Machine Learning community in both theoretical and applied algorithmical ways. In this paper, we demonstrate the strength of the algebraic combinatorial approach by applying it to the basic theoretical questions in matrix completion. We derive necessary and sufficient conditions for when a matrix of arbitrary rank can be reconstructed from a set of entries. As special cases, we derive central results for matrix completion in terms of these entries, which can be seen as analoga of results in rigidity theory. We also give a new algorithm, inspired by the newly derived criteria, which can be used to reconstruct matrices where the state-of-the-art algorithms may fail. We conclude with numerical simulations, comparing the novel criteria and matrix completion algorithm to state-of-the-art methods.

Summing up, our novel contributions are:

- Introduction of a combinatorial algebraic framework for the problem of matrix completion, building on known algebraic and combinatorial techniques from determinantal varieties and rigidity theory

- The first necessary and sufficient identifiability conditions for arbitrary rank matrices in terms of the set of known entries

- Formulation of a constructive combinatorial algebraic algorithm for matrix completion

Before we continue with stating the results, we want to give a brief overview of the main ideas of the paper. Suppose the truth, or generative model, is given by a $3 \times 3$ matrix, say

$$A = \begin{pmatrix} a_{11} & a_{21} & a_{31} \\ a_{12} & a_{22} & a_{32} \\ a_{13} & a_{23} & a_{33} \end{pmatrix} = \begin{pmatrix} 1 & 2 & 3 \\ 2 & 4 & 6 \\ 4 & 8 & 12 \end{pmatrix}.$$

The entries $a_{ij}$ may be interpreted as class or index dependent observations. The matrix $M$ has rank 1, so the entries are highly dependent; for example, one single entry in a row can be used to predict the whole row if an arbitrary other row is known. In the matrix completion setting, only some of the entries are known - the so-called measurements. Suppose we make five measurements. Two possible scenarios are that we measure the entries in

$$A_1 = \begin{pmatrix} 1 & 2 & 3 \\ * & 4 & * \\ 4 & * & * \end{pmatrix} \quad \text{or in} \quad A_2 = \begin{pmatrix} 1 & * & 3 \\ * & 4 & * \\ 4 & * & 12 \end{pmatrix}.$$

If we know that the original matrix $A$ had rank one, we can reconstruct $A$ from $A_1$. Namely, the column space is one-dimensional, so the entries of each column are determined by a unique scaling factor. The scaling factor can be determined for each of the two missing columns in $A_1$ by computing $a_{22}/a_{21}$, resp. $a_{31}/a_{11}$. Algebraically, this corresponds to successively solving all minor equations

$$a_{ij}a_{k\ell} = a_{i\ell}a_{kj},$$

or, as above, expressed as fractions, the equations

$$\frac{a_{ij}}{a_{i\ell}} = \frac{a_{kj}}{a_{k\ell}}.$$

If we try this approach to reconstruct $A$ from $A_2$, it fails. Indeed, it turns out that it is impossible to reconstruct $A$ from $A_2$. One possible way to see this is that the entry $a_{12}$ can be chosen independently to be anything, and then continuing the strategy for $A_1$ gives a valid matrix completion for any choice of $a_{12}$. Thus, the set of valid completions is highly non-unique - it has one degree of freedom - or, algebraically spoken, (Krull) dimension one.

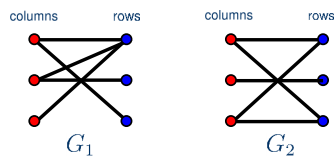

Alternatively, the problem can be regarded combinatorially. To both the matrices $A_1, A_2$, we can associate bipartite graphs $G_1, G_2$ in the following way: the bipartite vertex sets are the columns ("red vertices") and the rows ("blue vertices") of $A_k$. An edge $(i, j)$ is in $G_k$ if the entry $a_{ij}$ is known in $A_k$, i.e. $A_k$ has no $*$ at that position. Above is a depiction of $G_1$ and $G_2$. One can see that $G_1$ is connected, while $G_2$ is not. Intuitively, the edges one can add by virtue of the minor equations correspond to "possibilities to reconstruct". One can show that one can add exactly the edges between vertices in the same connected component; thus, for a unique reconstruction of $A$, the graph must be connected.

This shows that matrix completion has intrinsically both algebraic and combinatorial features; the phenomena above have indeed interesting generalizations. In the following sections, we will give a more rigorous exposition which will lead to novel types of theoretical results and algorithms.

## 2. Theory of Matrix Completion

### 2.1. Matrix Completion - an Algebraic Problem

First we introduce some definitions and notations which allow us to formulate the problem of matrix completion and later enable us to state our results. The central observation is that the relevant occurring sets and maps in matrix completion are all algebraic (i.e. defined by polynomials) and thus give rise to an algebraic formulation of matrix completion. The general setting for matrix completion is the following: one starts with a low-rank $(m \times n)$ matrix $A$, say of rank $\operatorname{rank} A = r$. However, only a fixed subset of the entries of $A$ is known. The task is now to reconstruct $A$ uniquely from the known entries, and the fact that $A$ has rank $r$. Whether this is possible depends obviously on the known entries, and in this paper, we will investigate this question. We model the choice of entries as follows:

**Definition 2.1.** A map $\Omega : \mathbb{C}^{m \times n} \to \mathbb{C}^\alpha$ which sends a matrix to a fixed tuple of its entries, i.e.

$$\Omega : (a_{ij})_{\substack{1 \le i \le m \\ 1 \le j \le n}} \mapsto (a_{i_1 j_1}, a_{i_2 j_2}, \ldots, a_{i_\alpha j_\alpha}),$$

where the bituples $(i_\ell, j_\ell)$ are all different, is called *masking*. Such a map is uniquely defined by the set of entries $i_k j_k$ in the image set. We call the unique matrix which has ones at those entries, and zeroes elsewhere, the *mask* of $\Omega$ and denote it by $M(\Omega)$.

Note that a masking $\Omega$ can be interpreted as taking the Hadamard product of the argument with the mask $M(\Omega)$. Matrix completion requires one to restrict maskings to the set of low rank matrices, for which we introduce notation:

**Notations 2.2.** We will denote the set of $(m \times n)$-matrices with rank at most $r$ as

$$\mathcal{M}(r; m \times n).$$

In the following, we assume $r \le m \le n$ without loss of generality. A classical result in algebraic geometry states that $\mathcal{M}(r; m \times n)$ is an algebraic variety, the so-called *determinantal variety* (of low-rank matrices); that is, the set $\mathcal{M}(r; m \times n)$ is a closed subset of $\mathbb{C}^{m \times n}$, defined by the vanishing of a set of polynomials - namely, the $(r+1)$-minors. The variety $\mathcal{M}(r; m \times n)$ is known to be irreducible, i.e., it cannot be written as the proper union of two algebraic varieties. Moreover, the dimension of the determinantal variety is known to be

$$\dim \mathcal{M}(r; m \times n) = (m + n - r) \cdot r.$$

With these notations, the identifiability problem of matrix completion can be reformulated as follows:

**Problem 2.3.** Let $\Omega$ be a masking. When is its restriction to the (at most) rank $r$ matrices

$$\Omega : \mathcal{M}(r; m \times n) \longrightarrow \mathbb{C}^\alpha$$

*invertible? That is, when is it uniquely possible to identify a matrix $A \in \mathcal{M}(r; m \times n)$ from its masking $\Omega(A)$, depending on $m, n, r$ and $M(\Omega)$? Alternatively formulated: when does it hold that*

$$\{A\} = \Omega^{-1}(\Omega(A))$$

Note that in this formulation, invertibility may also depend a-priori on the choice of the matrix $A$. We will synonymously use the term *completability* when $A$ can be uniquely identified. One might now naively think that invertibility does not depend on the choice of $A$; however, the following result shows that this is not the case in general:

**Proposition 2.4.** *Let $r \ge 2$. Then the restricted masking $\Omega : \mathcal{M}(r; m \times n) \longrightarrow \mathbb{C}^\alpha$ is injective if and only if $\alpha = mn$.*

*Proof.* Clearly, if $\alpha = mn$; then $\Omega$ is injective, as it is the identity map. So it suffices to prove: if $r \ge 2$ and $\alpha < mn$, there exists a matrix $A$ such that $\{A\} \ne \Omega^{-1}(\Omega(A))$. Now since $\alpha < mn$, there exists an index $ij$ such that $M(\Omega)_{ij} = 0$. Let $A$ be any matrix whose columns, except the $j$-th, span an $(r-1)$-dimensional vector space. Since $X$ is of (at most) rank $r$, the set $\Omega^{-1}(\Omega(A))$ contains any matrix $\tilde{A}$ which is identical to $A$ but has an arbitrary entry at the index $ij$. □

That, in general, not all matrices can be reconstructed is a known result, see (Candes & Recht, 2009); indeed,

Proposition 2.4 may seem to imply that we have to look for additional properties or sampling assumptions on the possible $A$ to obtain firm results - which is the rationale behind introducing spikiness or incoherence. However, the pathologies presented in the proof can be remedied by a slight relaxation to all matrices but a zero set, which is formally modelled by the "generic" property:

**Definition 2.5.** A *generic* $(m \times n)$-matrix of rank $r$ is a positive, continuous, $\mathcal{M}(r; m \times n)$-valued random variable. By convention, we will say that a statement is true for a generic matrix, if it is true with probability one.

Note that in the definition, we have assumed nothing on the distribution of the random variable. In fact, a statement is true with probability one for one particular positive continuous distribution if and only if it is true with probability one for all such distributions, as being true generically is equivalent that the exception set to the statement is a zero set in $\mathcal{M}(r; m \times n)$. Moreover, there is a purely algebraic concept of "generic" in algebraic geometry, which is - under certain conditions - equivalent to our probabilistic formulation. Details can be found e.g. in the appendix of (Király et al., 2012a). Under the slight relaxation provided by genericity, it can be proved that the completability of a matrix depends only on the chosen mask (Recall that our convention in Definition 2.5 allows for a possible zero set of exceptional matrices):

**Theorem 2.6.** Let $A$ be a generic $(m \times n)$ matrix of rank $r$. Then the dimension of the algebraic set $\Omega^{-1}(\Omega(A))$ depends only on the mask $M(\Omega)$. Moreover, if the dimension is zero, that is, the number of possible completions in $\Omega^{-1}(\Omega(A))$ is finite, then this number also depends only on the mask $M(\Omega)$. In particular, the completability of the matrix $A$ depends only on the chosen mask.

*Proof.* As $A$ is a positive continuous random variable, the statement above is equivalent to showing that there is an open dense set $U$ in $\mathbb{C}^{m \times n}$, such that for all matrices $B$ in $U$, the set of possible completions $\Omega^{-1}(\Omega(B))$ has the same dimension and cardinality. But this is exactly what is stated in the generic fiber theorem or upper semicontinuity theorem (see e.g. I.8, Corollary 3 in (Mumford, 1999)), when applied to the algebraic map $\Omega : \mathcal{M}(r; m \times n) \longrightarrow \mathbb{C}^\alpha$, considering that $\mathcal{M}(r; m \times n)$ is irreducible. □

Theorem 2.6 motivates the following relaxations of injectivity and finiteness:

**Definition 2.7.** A masking $\Omega$ is called *generically injective* on $\mathcal{M}(r; m \times n)$ if for generic $A$, one has $\{A\} = \Omega^{-1}(\Omega(A))$. A masking $\Omega$ is called *generically finite* on $\mathcal{M}(r; m \times n)$ if for generic $A$, the set $\{A\} = \Omega^{-1}(\Omega(A))$ is finite. Abbreviatingly, we will also say that $\Omega$ is generically injective (or finite) in rank $r$ if $m, n$ is arbitrary.

Note that generic finiteness is a necessary condition for generic injectivity. With the result of Theorem 2.6 and this notation, we arrive at the definitive formulation of the identifiability of matrix completion:

**Problem 2.8.** *When is a masking $\Omega$ generically injective on $\mathcal{M}(r; m \times n)$, given $m, n, r$ and $M(\Omega)$?*

We will investigate this question in greater detail in the next section. Also note that a generic matrix in $\mathcal{M}(r; m \times n)$ always has the maximal rank $r$ (i.e., with probability one). Before continuing, we mention an important detail:

**Remark 2.9.** A necessary condition for $\Omega$ to be generically injective is that it is generically finite. The latter is, by Definition 2.7, equivalent to the fact that

$$\dim \Omega^{-1}(\Omega(A)) = 0$$

for a generic matrix $A \in \mathcal{M}(r; m \times n)$. A classical result in algebraic geometry relates the generic fiber dimension to the dimensions of range and image: it equates to

$$\dim \Omega^{-1}(\Omega(A))$$
$$= \dim \mathcal{M}(r; m \times n) - \dim \Omega(\mathcal{M}(r; m \times n))),$$

as the determinantal variety $\mathcal{M}(r; m \times n)$ is irreducible; see e.g. I.8, Theorem 3 in (Mumford, 1999) for a proof. The generic fiber dimension is also the same as the relative tangent space dimension at the generic point of the map $\Omega$. For fixed $\Omega$, this dimension can be calculated by determining the rank (or co-rank) of the Jacobi matrix over a randomly chosen $A$ - this is a classical task in computational algebra. In (Singer & Cucuringu, 2010), a probabilistic algorithm along these lines is provided for the rigid realization problem - the vanishing of the dimension is called "local completability"; the same techniques can be used to construct an algorithm for testing generic injectivity, or generic finiteness of a mask. A more detailed discussion of the relation between rigid realization and matrix completion can be found in the supplement (Király & Tomioka, 2012).

### 2.2. Matrix Completion - a Combinatorial Problem

We continue by deriving combinatorial conditions for a mask to be completable. We will show that completability is encoded in the combinatorial properties of a specific graph:

**Definition 2.10.** Let $\Omega$ be a masking with mask $M(\Omega)$. We will call the unique bipartite graph $G(\Omega)$ which has adjacency matrix $M(\Omega)$ the *adjacency graph* of $\Omega$. Recall that to the $(m \times n)$-matrix $M(\Omega)$, the associated bipartite graph $G(\Omega) = (V_1, V_2, E)$ is defined as follows: The two sets of vertices are the numbers $V_1 = \{1, \ldots, m\}$ and $V_2 = \{1, \ldots, n\}$; an edge $(i,j) \in V_1 \times V_2$ is in $E$ if and only if $M(\Omega)_{ij} = 1$. We will denote the above edge set $E$ of $G(\Omega)$ by $E(\Omega)$. In the following, for convenience of reading, we will refer to the elements of $V_1$ as the *red*, and to the elements of $V_2$ as the *blue* vertices of $G(\Omega)$.

A sufficient condition for matrix completability can be formulated in terms of a generalized closure:

**Definition 2.11.** Let $G$ be a bipartite graph. Let us denote by $K_{r+1,r+1}^-$ the complete bipartite graph $K_{r+1,r+1}$ minus one edge[1]. If $G$ does not contain a vertex set whose induced subgraph is isomorphic to $K_{r+1,r+1}^-$, we call $G$ an *r-closed* bipartite graph, and $G$ is its *r-closure*. If $G$ contains a vertex set whose induced subgraph is isomorphic to $K_{r+1,r+1}^-$, denote by $G'$ the graph which is obtained from $G$ by adding the missing edge in $G$. In this case, the $r$-closure of $G$ is recursively defined as the $r$-closure of $G'$. If the $r$-closure of a bipartite graph $G$ with $m$ red and $n$ blue edges is $K_{m,n}$, we call $G$ an *r-closable* graph.

The recursive definition of $r$-closure is well-defined, as the possible number of edges in $G$ is finite, and by adding edges one arrives always at the same unique $r$-closed graph. Also, the 1-closure of a bipartite graph is its bipartite transitive closure. One can show that a bipartite graph is connected if and only if it is 1-closable.

In this terminology, we may formulate a necessary condition for generic injectivity:

**Proposition 2.12.** *A masking $\Omega$ is generically injective in rank $r$ if $G(\Omega)$ is $r$-closable.*

*Proof.* The proof uses that $r$-closability is the combinatorial counterpart of being able to subsequently find $(r+1) \times (r+1)$-blocks in the mask containing exactly one zero. For each of these blocks, the missing entry can be calculated using the vanishing minor condition in the generic situation. Details can be found in the supplement (Király & Tomioka, 2012). □

Similarly, one can derive necessary conditions for generic injectivity by combining combinatorial and algebraic considerations:

[1] $K_{r+1,r+1}$ has $r+1$ red and $r+1$ blue vertices, plus all possible edges.

**Proposition 2.13.** *A masking $\Omega$ is generically injective, or generically finite, on $\mathcal{M}(r; m \times n)$ only if:*

(i) $\#E(\Omega) \geq r \cdot (m + n - r)$

(ii) *Each vertex of $G(\Omega)$ has degree at least $r$*

(iii) *$G(\Omega)$ is $r$-connected, i.e., $G(\Omega)$ is connected after removing an arbitrary set of $r$-1 edges.*

(iv) *For a vertex partition of $G(\Omega)$ into $N$ bipartite subgraphs $G_1, G_2, \ldots, G_N$ such that $G_i$ has $m_i$ red and $n_i$ blue vertices and edge set $E_i$, with $\sum_{i=1}^N m_i = m$ and $\sum_{i=1}^N n_i = n$, the number of edges in $G$ connecting vertices in two different $G_i$ is at least*

$$\left(\#E(\Omega) - \sum_{i=1}^N E_i\right) \geq r \cdot (m + n - r)$$
$$- \sum_{i=1}^N \left(m_i n_i - \max(0, m_i - r) \max(0, n_i - r)\right).$$

*Note that this number is at least $r$ for non-trivial partitions and at most $(N-1)r^2$.*

*Proof.* The proof follows the lines Remark 2.9 which states that $\Omega$ is generically injective only if the generic fiber dimension is zero, and provides a formula to obtain the generic fiber dimension from the dimension of the image and the range. This implicitly gives a way to bound the generic fiber dimension in terms of the number of known entries of the mask. The calculations and the proof are carried out in the supplement (Király & Tomioka, 2012). □

Interestingly, neither in the sufficient nor in the necessary condition is it used that the graph $G(\Omega)$ arises as the adjacency graph for a matrix completion problem. So, as sufficient conditions imply necessary ones, we may formulate a purely graph theoretic result:

**Theorem 2.14.** *Any $r$-closable bipartite graph is $r$-connected. Moreover, any $r$-closable bipartite graph $G$ with $m$ red and $n$ blue vertices fulfills conditions (i) to (iv) from Proposition 2.13.*

We conclude with a result which relates the necessary and sufficient condition in special cases:

**Proposition 2.15.** *For $r = 1$ and $r = m - 1$ (recall that we have assumed $r \leq m \leq n$), the necessary and sufficient conditions in Propositions 2.13 and 2.12 coincide. In particular, a 1-connected bipartite graph is 1-closable, and any bipartite graph fulfilling conditions (i) and (ii) from Proposition 2.13 with $r = m - 1$ is $(m-1)$-closable.*

*Proof.* The proof is carried out in detail in the supplement (Király & Tomioka, 2012). For $r = 1$, the statement follows from the purely graph-theoretical fact that 1-closability is equivalent to 1-connectedness - that implies the equivalence of the necessary condition (iii) in Proposition 2.13 and the sufficient condition from Proposition 2.12. For $r = m - 1$, the statement follows from a combinatorial pigeonhole-type argument which shows that conditions (i) and (ii) in Proposition 2.13 already imply $r$-closability. □

We want to mention that the result for $r = 1$ is the matrix completion analogue of Proposition 5.6 in (Singer & Cucuringu, 2010) for rigid realization; the result for $r = m - 1$ is similar to Corollaire 1.5 of (Giusti & Merle, 1982) which is for coordinate sections. Moreover, for $r \geq 2$, the conditions in Propositions 2.13 and 2.12 are not equivalent for generic injectivity, as the following example shows:

**Example 2.16.** Consider the case of rank $r = 2$, $m = n = 4$ and the masks

$$M_1 = \begin{pmatrix} 1 & 1 & 1 & 1 \\ 1 & 1 & 1 & 1 \\ 1 & 1 & 0 & 0 \\ 1 & 1 & 0 & 0 \end{pmatrix} \text{ and } M_2 = \begin{pmatrix} 0 & 1 & 1 & 1 \\ 1 & 0 & 1 & 1 \\ 1 & 1 & 0 & 1 \\ 1 & 1 & 1 & 0 \end{pmatrix}.$$

Both matrices $M_1$ and $M_2$ fulfill the necessary conditions given in Proposition 2.13. For example, Condition (i) is fulfilled, as both matrices contain $r \cdot (m+n-r) = 12$ entries. However, only $M_1$ fulfills the sufficient condition in Proposition 2.12: $M_1$ is 2-closable and thus defines a generically injective masking by Proposition 2.12 with unique completion; the matrix $M_2$ is not 2-closable. Indeed, for a generic matrix in $\mathcal{M}(2; 4 \times 4)$ masked by $M_2$, there exist two distinct completions, as an elementary calculation of the vanishing minor condition shows.

However, we cannot exclude that Propositions 2.13 is equivalent to generic finiteness, which - after adding one suitable edge - implies generic injectivity. So it may be that the necessary and sufficient conditions for generic injectivity are tight up to one edge.

### 2.3. An Algebraic Combinatorial Algorithm

The necessary and sufficient conditions given in Propositions 2.12 and 2.13 give theoretical bounds for the success of matrix completion. We wish to demonstrate that these conditions can be also used constructively to obtain novel matrix completion algorithms. First we describe an algorithm for noise-free entries with Algorithm 1. It is based on the sufficient condition in Proposition 2.12 which is essentially algorithmic and, with a slight modification, yields an explicit procedure that completes the matrix by explicitly calculating entries for the $r$-closure.

---
**Algorithm 1** Matrix completion by $r$-closure.
*Input:* A masked rank $r$ matrix $\Omega(A)$.
*Output:* The complete matrix $A$, or nil.

1: Optional: Check whether some necessary conditions in Proposition 2.13 hold; e.g. conditions (i) and (ii). If not, return nil.
2: Find a bipartite subgraph of $G(\Omega)$ isomorphic to $K^-_{r+1,r+1}$; find the corresponding $(r+1 \times r+1)$-sub-matrix of $A$, missing one element. If such a subgraph does not exist, return nil.
3: Compute the missing element of $A$ by (numerically stable) Gaussian elimination.
4: Add the missing element to $A$; add the entry to the mask of $\Omega$.
5: Repeat 2-4 until $G(\Omega) = K_{m,n}$.

---

Algorithm 1 computes a completion, whenever the sufficient conditions in Proposition 2.12 holds. If not, th algorithm will return nil. In line 1, the algorithm checks necessary conditions for completability. Alternatively, one could use the algorithm given in (Singer & Cucuringu, 2010) to obtain a definite answer - depending on available computational time. In lines 2-4, an entry is added to $A$ using the determinantal condition. The computationally critical step in this part is finding a subgraph isomorphic to $K_{r+1,r+1}$ minus one edge. While the subgraph problem is known to be computationally hard, heuristics (e.g. random sampling on a subset) can speed up this part considerably under the cost of correctness, or termination. However, if the algorithm terminates, the found completion is a valid solution. If the matrix is $r$-closable, repetition of 2-4 then leads to reconstructing the whole matrix. This algorithm for the noise-free case can be extended to cope with noise, using methods from approximate linear algebra and approximate commutative algebra as in (Király et al., 2012b). Due to space limitations, we refrain from doing so, and will present the results in a future paper.

As already noted, it cannot be a-priori excluded that Proposition 2.12 fails even if there is a finite completion; as Proposition 2.12 is not proven to be necessary. However, in our experiments, this seems to be never the case for complete subgraph search.

## 3. Experiments

In our simulations, we investigate the necessary and sufficient conditions from Propositions 2.12 and 2.13 in comparison to our new Algorithm 1.

The presented version of Algorithm 1 uses a heuristic to find the $r$-closable subgraphs of $G(\Omega)$: for each edge in the bipartite graph $G(\Omega)$, the vertex degree is determined. Then, the algorithm tries to construct closable subgraphs from the highest degree neighbor vertices. If this fails for a fixed number of trials, the algorithm goes to the next vertex.

We also include two state-of the-art algorithms in our comparison: the nuclear norm heuristic (Fazel et al., 2001; Srebro et al., 2005) and OptSpace (Keshavan & Oh, 2009; Keshavan et al., 2010). The nuclear norm heuristic solves the convex minimization problem

$$\text{minimize } \|A\|_* \text{ such that } \Omega(A) = \tilde{A},$$

where $\tilde{A}$ is the matrix of measurements. We use the alternating direction method of multipliers (Gabay & Mercier, 1976) to solve the above minimization problem. The second algorithm, OptSpace, solves the non-convex minimization problem

$$\text{minimize } \|\Omega(A) - \tilde{A}\| \text{ such that rank}(A) \leq r,$$

using gradient descent on the Grassmann manifold (Keshavan & Oh, 2009). We compare the necessary condition, and the algorithms in three experiments. For each experiment, we first fix the matrix size $m, n$ and the matrix rank $r$. Then, we vary the number of measurements, i.e. the number of known entries in $\Omega(A)$, or in previous notation, $\#E(\Omega)$. For each number of measurements, we randomly and uniformly sample $N = 100$ masks without replacement, and for each mask, a random $(m \times n)$ rank $r$ matrix $A$. We then calculate the ratio of masks for which

- The necessary condition of Propositions 2.13 (iii) holds, i.e. $M$ is $r$-connected[2].
- Algorithm 1 is able to compute the completion $A$, which implies that $M$ is $r$-closable.
- The OptSpace algorithm is able to compute the completion $A$
- The nuclear norm heuristic is able to compute the completion $A$

The results are shown in Figure 1. For small rank $r = 3$, Algorithm 1 outperforms the state-of the art methods for the small $(10 \times 15)$ as well as for the big $(40 \times 50)$-matrices. For $(40 \times 50)$-matrices with rank 10, $r$-closure is comparable to Nuclear Norm and outperforms OptSpace. In the cases with small rank, $r$-connectivity and $r$-closability seem to be close to each other, but become less close with increasing rank. In all cases, Algorithm 1 is at least competitive with the existing state-of-the-art methods.

---

[2]We used the max-flow min-cut duality to decide $r$-connectivity. More specifically, we solve $m+n-1$ max-flow problems to obtain the max flow between the first blue vertex and one of the remaining $m + n - 1$ vertices over the bipartite graph $G(\Omega)$ with unit edge weights. If the minimum over the $m + n - 1$ max flows is less than $r$, the bipartite graph $G(\Omega)$ is not $r$-connected, otherwise it is $r$-connected. The MatlabBGL library was used to compute max-flow.

## 4. Conclusion

In this paper, we have introduced a basic combinatorial algebra framework for the matrix completion problem. With this framework, we were able to derive the first known necessary and sufficient conditions for completability of a matrix in terms of the set of known entries, building on the introduced techniques which link matrix completion to combinatorial graph theory and algebraic geometry. As a by-product, we have also obtained graph theoretical results. Following the sufficient condition, we were also able to formulate an algorithm for the matrix completion problem, which we have shown to be competitive with known matrix completion algorithms and which performs better than state-of-the art methods for small rank. The obtained results lead us to argue that studying the interactions between machine learning, algebraic geometry, and combinatorics bears the potential to be highly beneficial for the problem of matrix completion, regarded from the viewpoint of any of the three fields. These links are not yet fully explored; however, the presented results suggest a fruitful interaction. We conclude that future work on matrix completion can only benefit from an interdisciplinary connection of machine learning and combinatorial algebra.

## Acknowledgements

We would like to thank Takeaki Uno and Louis Theran for valuable comments. This research was partially supported by MEXT KAKENHI 22700138 and the Global COE program "The Research and Training Center for New Development in Mathematics".

# A Combinatorial Algebraic Approach for the Identifiability of Low-Rank Matrix Completion

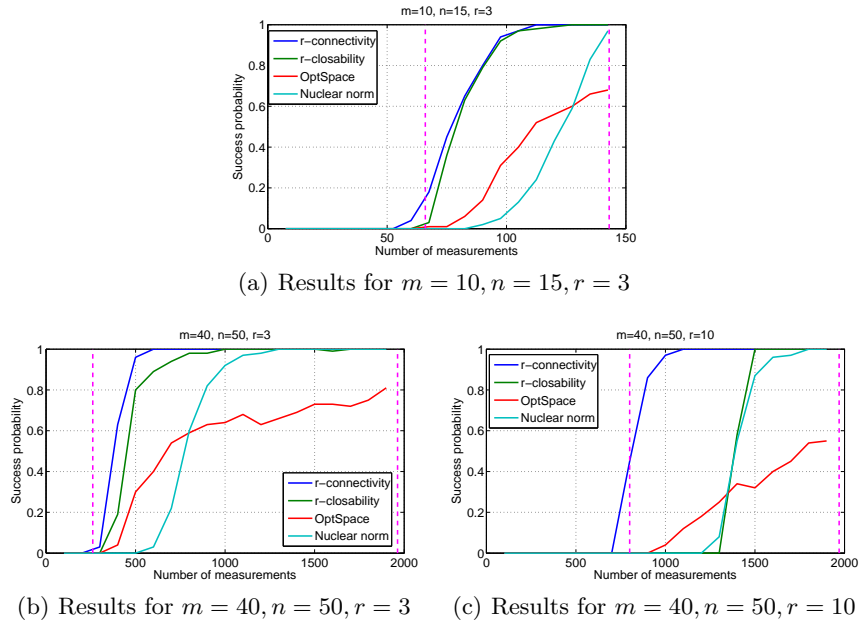

(a) Results for $m = 10, n = 15, r = 3$

(b) Results for $m = 40, n = 50, r = 3$

(c) Results for $m = 40, n = 50, r = 10$

*Figure 1.* For three different choices of matrix size $m, n$ and matrix rank $r$, the figure shows, for different numbers of random measurements, the estimated success rates of: the necessary condition in Propositions 2.13 (iii), r-connectivity; a heuristic implementation of Algorithm 1, which implies $r$-closability if successful; the OptSpace algorithm; the nuclear norm algorithm. Success rates are estimated over $N = 100$ random masks with fixed number of random measurements. The left dotted line is the necessary condition in Propositions 2.13 (i), at $r \cdot (m + n - r)$ entries, the right dotted line is at $m(n-1) + r$ entries where all matrices are identifiable. We would like to note that in each of the three experiments, the curve for 2.13 (ii) would lie over (iii) when plotted, while being a strictly weaker condition.